\documentclass[letterpaper]{article} 
\usepackage{aaai24}  
\usepackage{times}  
\usepackage{helvet}  
\usepackage{courier}  
\usepackage[hyphens]{url}  
\usepackage{graphicx} 
\usepackage{natbib}  
\usepackage{caption} 
\usepackage{algorithm}
\usepackage{algorithmic}
\usepackage{amsmath}
\usepackage{amssymb}
\usepackage{multirow}
\usepackage{array}
\usepackage{xcolor}
\usepackage{subcaption}
\usepackage{newfloat}
\usepackage{listings}
\DeclareCaptionStyle{ruled}{labelfont=normalfont,labelsep=colon,strut=off} 
\floatstyle{ruled}
\newfloat{listing}{tb}{lst}{}
\floatname{listing}{Listing}
\title{UCMCTrack: Multi-Object Tracking with Uniform Camera Motion Compensation}
\author{
    Kefu Yi\textsuperscript{\rm 1}\thanks{Corresponding author.}, Kai Luo\textsuperscript{\rm 2}, Xiaolei Luo\textsuperscript{\rm 2}, Jiangui Huang\textsuperscript{\rm 2}, Hao Wu\textsuperscript{\rm 2}, Rongdong Hu\textsuperscript{\rm 3}, Wei Hao\textsuperscript{\rm 1}\footnotemark[1]\\
}
\affiliations{
    \textsuperscript{\rm 1}School of Traffic and Transportation,
Changsha University of Science and Technology, Changsha, China\\
    \textsuperscript{\rm 2}College of Automotive and Mechanical Engineering, Changsha University of Science and Technology, Changsha, China\\
    \textsuperscript{\rm 3}Changsha Intelligent Driving Institute, Changsha, China\\
    
    \{corfyi, haowei\}@csust.edu.cn
}

\usepackage{bibentry}

\begin{document}

\maketitle

\begin{abstract}
Multi-object tracking (MOT) in video sequences remains a challenging task, especially in scenarios with significant camera movements. This is because targets can drift considerably on the image plane, leading to erroneous tracking outcomes. Addressing such challenges typically requires supplementary appearance cues or Camera Motion Compensation (CMC). While these strategies are effective, they also introduce a considerable computational burden, posing challenges for real-time MOT. In response to this, we introduce UCMCTrack, a novel motion model-based tracker robust to camera movements. Unlike conventional CMC that computes compensation parameters frame-by-frame, UCMCTrack consistently applies the same compensation parameters throughout a video sequence. It employs a Kalman filter on the ground plane and introduces the Mapped Mahalanobis Distance (MMD) as an alternative to the traditional Intersection over Union (IoU) distance measure. By leveraging projected probability distributions on the ground plane, our approach efficiently captures motion patterns and adeptly manages uncertainties introduced by homography projections. Remarkably, UCMCTrack, relying solely on motion cues, achieves state-of-the-art performance across a variety of challenging datasets, including MOT17, MOT20, DanceTrack and KITTI. More details and code are available at \url{https://github.com/corfyi/UCMCTrack}.
\end{abstract}

\begin{figure}[t]
\centering
\includegraphics[width=1.0\columnwidth]{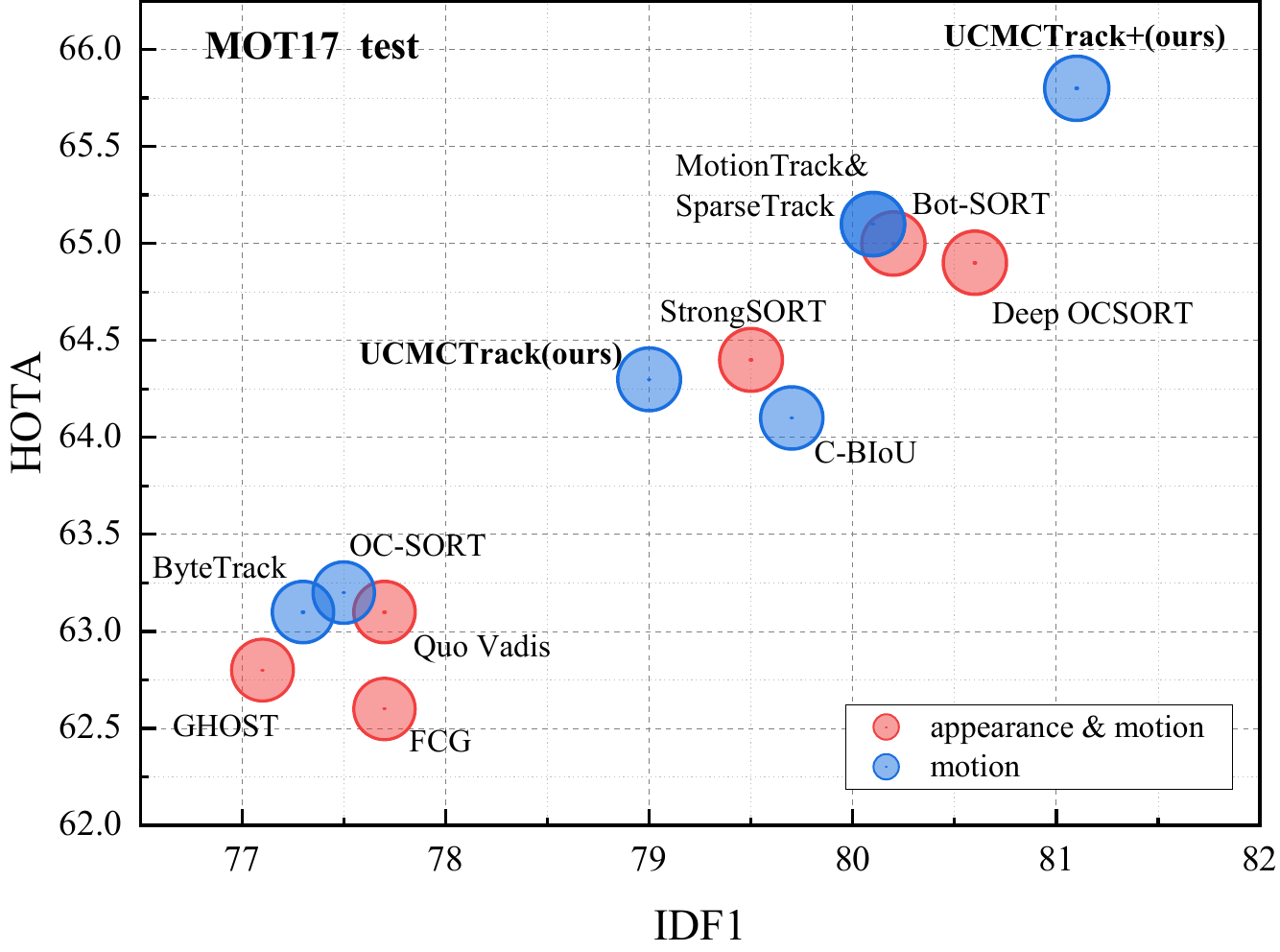} 
\caption{IDF1-HOTA-AssA comparisons of different trackers on
the test set of MOT17. The horizontal axis is IDF1,the vertical axis is HOTA, and the radius of circle is AssA. Our UCMCTrack+ achieves 65.8 HOTA, 81.1 IDF1 on MOT17 test, possessing significant competitiveness compared to SOTA trackers. Details are given in Table \ref{table1}.}
\label{fig1}
\end{figure}

\section{Introduction}

At the core of tracking-by-detection paradigm of multi-object tracking (MOT) is the accurate association of detections with tracked objects. Motion cues are widely used due to their effectiveness and simplicity. However, the application of motion model in scenarios with frequent camera movement is highly challenging.  This issue is usually addressed by applying additional appearance cues or performing frame-by-frame Camera Motion Compensation (CMC) on video captured by the moving camera. While effective, these additional measures introduce a non-negligible computational burden, posing a obstacle for real-time MOT. 

Thus, a pertinent question arises: Is it possible to employ motion cues in MOT that are robust to camera movement without resorting to the cumbersome frame-by-frame CMC? \textbf{Our answer is YES.} We have developed a pure motion model-based multi-object tracker that is robust to camera movement. For the same video sequence, it suffices to use the same camera motion compensation  parameters, rather than computing the camera motion compensation parameters for every frame as traditional CMC does. We choose to model the target's motion using a simple Kalman filter on the ground plane, instead of on the imaging plane as most MOT algorithms do, and effectively compensate the motion estimation errors caused by camera movement through the process noise parameters of the Kalman filter. We abandon the commonly used Intersection over Union (IoU) , and instead propose the Mapped Mahalanobis Distance (MMD). It computes the projected probability distribution on the ground plane, and utilizes the Mahalanobis distance to calculate the matching costs between targets. It not only effectively leverages the underlying motion patterns of the targets on the ground plane but also efficiently handles the uncertainties caused by homography projection.

A deep dive into motion-based MOT highlights a significant challenge when employing motion cues in highly dynamic scenes. Historically, IoU has been the favored metric for data association. On the surface, employing IoU on the image plane appears to be a more direct approach. However, its application often leads to inaccurate tracking outcomes, particularly in complex scenes marked by frequent camera movements. Notably, in these settings, detection and tracking boxes might completely fail to overlap, as shown in Figure \ref{fig3}. This observation underscores an imminent necessity: a transition from exclusive reliance on the image plane to harnessing the more robust motion patterns inherent to the ground plane. Embracing such a paradigm shift stands to effectively address challenges spawned by camera movements, setting the stage for superior tracking accuracy. Distinct from the vast majority of trackers relying IoU on the image plane, ground plane-based association can effectively considers camera movement as noise within the motion model. It minimizes the problems induced by camera movements. This methodology is notably more direct, convenient, and efficient than compensating for camera motion frame-by-frame via traditional CMC.

In light of these challenges, we introduce the Uniform Camera Motion Compensation (UCMC) tracker. It is a pure motion-based multi-object tracker that offers a holistic solution robust to camera jitter and motion, without any dependency on IoU-based methodologies. 

The main contributions of this paper are threefold:

\begin{itemize}
\item In the realm of multi-object tracking where IoU is conventionally employed to capitalize on motion cues, our work introduces an innovative non-IoU distance measure, singularly driven by motion cues, and manifests state-of-the-art performance across multiple established datasets, marking a significant departure from traditional tracking techniques.
\item In addressing the challenge of camera movements, we propose a method that diverges from conventional camera motion compensation techniques. Instead of computing camera compensation parameters frame-by-frame for video sequences, our approach uniformly applies the same compensation parameters across the entire sequence, substantially reducing the computational burden typically associated with camera motion adjustments.
\item We introduce UCMCTrack, a simplistic yet efficacious multi-object tracker that employs a novel, standalone motion-based distance measure. This new measure has the potential to complement commonly-used distance metrics such as IoU and ReID. Remarkably, when provided with detections, UCMCTrack operates at a very fast speed, exceeding 1000 FPS using just a single CPU.
\end{itemize}

\begin{figure*}[t]
\centering
\includegraphics[width=2.1\columnwidth]{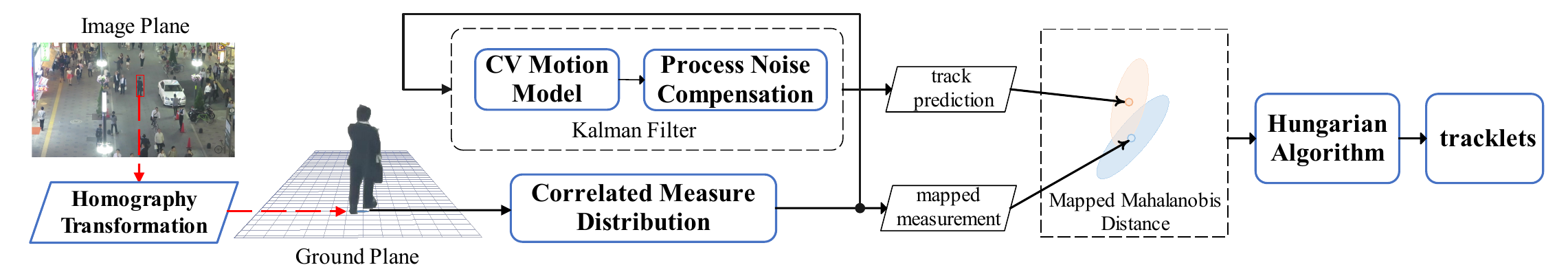} 
\caption{The pipeline of the proposed UCMCTrack. }
\label{fig2}
\end{figure*}

\section{Related Work}
\subsection{Distance Measures}

Distance measures play key roles in MOT to associate targets in the current frame with those in previous frames. Currently, most algorithms employ the pixel-based Intersection over Union (IoU) technique \cite{du2023strongsort,liu2020gsm}, which calculates the intersecting area between the detection box and the tracking box for target matching. However, in cases of camera jitter or low sampling rates, the two boxes may not intersect, rendering IoU ineffective. In contrast, Generalized Intersection over Union (GIoU) \cite{rezatofighi2019generalized} not only focuses on the overlapping region but also considers the non-overlapping area, thus improving the computation of image overlap. Distance-IoU (DIoU) builds upon GIoU by further incorporating geometric distance into the calculation \cite{zheng2020distance}. However, both these methods do not adequately represent the similarity in aspect ratios of the objects. Bidirectional Intersection over Union (BIoU) and the Cascade-BIoU (C-BIoU) proposed to augment the IoU-based approach by introducing a linear average motion estimation model and expanding the search region \cite{yang2023hard}. Nevertheless, all these methods operate in the image plane and cannot fully capture the actual motion patterns, leading to faulty tracking during camera motion. Recently, some methods have considered distance measures based on the ground plane. SparseTrack \cite{liu2023sparsetrack} goes beyond IoU and incorporates additional estimated pseudo-depth for supplementary metrics. Quo Vadis \cite{dendorfer2022quo} employs homography transformation to calculate the Euclidean distance in the bird's-eye view and combines it with IoU for target matching. Although these approaches utilize additional depth information, they still rely on IoU and fail to account for the uncertainty in the projection of targets onto the ground plane.

\subsection{Motion Models}

Tracking-by-detection MOT algorithms \cite{wojkeSimpleOnlineRealtime2017a,cao2022observation,maggiolino2023deep} often favor motion models for their simplicity and effectiveness. Among these, the Constant Velocity (CV) model, which assumes unvarying target motion between frames, is the most favored approach \cite{bewleySimpleOnlineRealtime2016,zhang2022bytetrack}. Numerous studies have been dedicated to improving motion estimation accuracy, employing methods such as Kalman filtering \cite{bewleySimpleOnlineRealtime2016,zhang2022bytetrack,zhou2020tracking}, optical flow \cite{xiao2018simple}, and displacement regression \cite{feichtenhofer2017detect,held2016learning}. However, current MOT algorithms \cite{Du_2021_ICCV,du2023strongsort,aharon2022bot} model the motion of tracking targets directly upon the image plane using detected bounding boxes. This approach fails to reflect the actual motion patterns of the targets on the ground plane, leading to erroneous tracking results during camera motion.

To further leverage the inherent motion patterns of the tracking targets, researchers \cite{LIU2020289,Marinello_2022_CVPR} have employed LSTM networks to predict target motion, while others \cite{BABAEE201969} have used RNN networks for similar purposes. Additionally, transformer networks \cite{yang2022transformer} have also been utilized to capture object motion patterns. 
In contrast to employing neural networks to explicitly predict target motion, the tracking-by-query propagation \cite{Zhang_2023_CVPR} forces each query to recall the same instance across different frames. Alternatively, the approach based on a Graphs framework \cite{Cetintas_2023_CVPR} is used to model data association. These methods use a learned network to implicitly grasp the dynamics of target motion. While they achieve promising results, their training process can be challenging, requiring a substantial amount of annotated data and computational resources. Moreover, complex network designs may not meet the real-time requirements on end devices.

\subsection{Camera Motion Compensation}

Camera Motion Compensation (CMC) is a prevalent method to address dynamic scenes in the field of MOT \cite{Bergmann_2019_ICCV,HAN202275,Khurana_2021_ICCV}. This is often achieved by aligning frames through image registration, leveraging techniques such as Enhanced Correlation Coefficient (ECC) maximization \cite{evangelidis2008parametric}, or employing feature matching methods like ORB \cite{rublee2011orb}. In BOT-SORT \cite{aharon2022bot}, image key-points were extracted frame-by-frame, with sparse optical flow subsequently applied. 
The affine transformation matrix of background motion is calculated and obtained via RANSAC \cite{fischler1981random}, and the affine matrix is used to transform the prediction box from the (k-1)-th frame coordinate system to the k-th frame coordinate system. In \cite{yu2019moving}, the pyramidal Lucas-Kanade optical flow is implemented to trace grid-based feature points. 
The affine transformation matrix between two consecutive frames is calculated through matching feature points, and the initial two frames and the background model are aligned with the current frame. In \cite{yeh2017three}, a camera motion compensation framework is proposed with utilization of temporal and spatial structure, which depends on pre-provided background model for background elimination, thereby posing challenges for its adaptation to new scenarios. However, when confronted with high-resolution videos, current CMC techniques impose substantial computational overhead and hinders the implementation of real-time target tracking.

\begin{figure*}[htbp]
  \centering
  \begin{subfigure}[b]{0.33\textwidth}
    \centering
    \includegraphics[width=\textwidth]{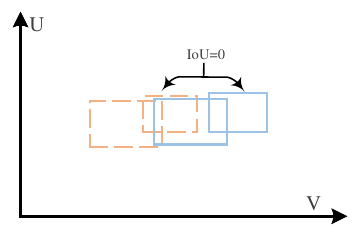}
    \caption{Image Plane (with IoU)}
    \label{Correlation_a}
  \end{subfigure}
  \begin{subfigure}[b]{0.33\textwidth}
    \centering
    \includegraphics[width=\textwidth]{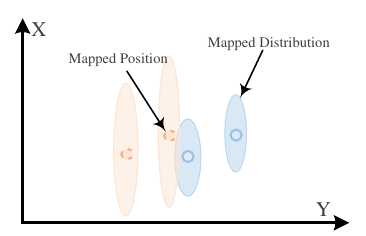}
    \caption{MMD (without CMD)}
    \label{Correlation_b}
  \end{subfigure}
   \begin{subfigure}[b]{0.33\textwidth}
    \centering
    \includegraphics[width=\textwidth]{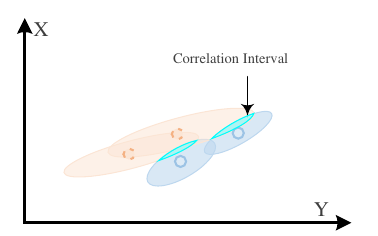}
    \caption{MMD (with CMD)}
    \label{Correlation_c}
  \end{subfigure}
  \caption{Visualization of distance measures. (a) Visualization of IoU on the image plane. IoU fails as there is no intersection between bounding boxes. (b) Visualization of Mapped Mahalanobis Distance (MMD) without Correlated Measurement Distribution (CMD). Incorrect associations occur due to insufficient utilization of distribution information. (c) Visualization of MMD with CMD. Correct associations after using the correlated probability distribution, undergoing a rotation on the ground plane.}
  \label{fig3}
\end{figure*}

\section{Method}

UCMCTrack follows the tracking-by-detection paradigm, with its pipeline detailed in Figure 2. We introduce significant advancements across crucial dimensions: motion model, distance measure, and process noise compensation. Together, these improvements bolster UCMCTrack, endowing it with  adaptability and efficiency across diverse tracking challenges. For the pseudocode please refer to Appendix A.

\subsection{Motion Modeling on Ground Plane}
We model objects' motion on the ground plane to better capture the fundamental essence of their motion patterns.
Selecting the appropriate state vector $\mathbf{x}$, observation vector $\mathbf{z}$, and determining the process noise $\mathbf{Q_k}$ and measurement noise $\mathbf{R}_{k}$ are crucial in establishing the Kalman constant velocity motion model.
In order to make observation and calculation more convenient, the choice was made to use the midpoint coordinates of the bottom edge of the bounding box in the image plane, projected onto the ground plane coordinates $x$ and $y$, as the observation vector. The state vector $\mathbf{x}$ is defined as $\mathbf{x} = [x, \dot{x}, y, \dot{y}]^\top$. According to the linear camera model \cite{yu2004general}, the mapping relationship between the ground plane coordinates $x$ and $y$, and the image plane coordinates $u$ and $v$, can be expressed as:

\begin{align}
\label{eq1}
        \left[ {\begin{array}{*{20}{c}}
        u\\
        v\\
        1
        \end{array}} \right]
        =\mathbf{A} \frac{1}{\gamma }\left[ {\begin{array}{*{20}{c}}
        {{x}}\\
        {{y}}\\
        1
        \end{array}} \right]
\end{align}
Where $\gamma$ is the scale factor, and $\mathbf{A}$ represents the projection matrix which is the product of camera intrinsic and extrinsic parameters. Please refer to Appendix B for more details.

\subsection{Correlated Measurement Distribution}

In general, the measurement errors of detectors on the image plane follow an independent normal distribution, and their covariance matrix $\mathbf{R}^{uv} _{k}$ can be represented as:
\begin{align}
\label{eq2}
    \mathbf{R}^{uv} _{k}  = \left[ {\begin{array}{*{20}{c}}
    {(\sigma_{m} {w}_{k})^2}&{0}\\
    {0}&{(\sigma_{m} {h}_{k})^2}\end{array}
    } \right]
\end{align}
Where: $\sigma_{m}$ represents the detection noise factor as a hyperparameter, and ${w}_{k}$ and ${h}_{k}$ denote the detected width and height from the detector.

If we express the inverse matrix of $\mathbf{A}$ in Eq. 1 as:
\begin{align}
\label{eq3}
    {\mathbf{A}^{ - 1}} = \left[ {\begin{array}{*{20}{c}}
    {{a_{11}}}&{{a_{12}}}&{{a_{13}}}\\
    {{a_{21}}}&{{a_{22}}}&{{a_{23}}}\\
    {{a_{31}}}&{{a_{32}}}&{{a_{33}}}
    \end{array}} \right]
\end{align}

This leads to the covariance matrix of measurement errors in the ground plane as: (Please refer to Appendix B for a detailed derivation)
\begin{align}
\label{eq4}
\mathbf{R}^{}_{k}= \mathbf{C}{\mathbf{R}^{uv}_{k} }{\mathbf{C}^T}
\end{align}
where:
\begin{align}
\label{eq5}
    \mathbf{C} 
    =\left[ {\begin{array}{*{20}{c}}
    {\gamma {a_{11}} - {a_{31}}{\gamma }{x}}&{\gamma {a_{12}} - {a_{32}}{\gamma}{x}}\\
    {\gamma {a_{21}} - {a_{31}}{\gamma }{y}}&{\gamma {a_{22}} - {a_{32}}{\gamma }{y}}
    \end{array}} \right]
\end{align}
Thus, we obtain the mapped measurement noise matrix $\mathbf{R}_k$ in the ground plane.  It's important to highlight that the mapped distribution exhibits a strong correlation since $\mathbf{R}_k$  is non-diagonal. This allows for more accurate association of targets on the ground plane.

\subsection{Mapped Mahalanobis Distance}

In image plane motion modeling, IoU is the most commonly used distance measure for data association. However, when objects are in high-speed motion or captured at low FPS or in scenes with moving camera, the lack of overlap between detection boxes and tracklets renders IoU ineffective. Conversely, by employing normalized Mahalanobis distance in ground plane modeling, the issue of IoU inefficiency is effectively addressed, as depicted in Figure \ref{fig3}.

The calculation of mapped Mahalanobis distance between track state and measurement involves three steps:

\begin{enumerate}
\item Calculate Residual:
\begin{align}
\label{eq6}
\mathbf{\epsilon} &= \mathbf{z} - \mathbf{H} \hat{\mathbf{x}}
\end{align}
   Here, $\mathbf{z}$ is the mapped measurement on ground plane, $\hat{\mathbf{x}}$ is the predicted track state, and $\mathbf{H}$ is the observation matrix.
\item Compute Residual Covariance Matrix:
\begin{align}
\label{eq7}
\mathbf{S} &= \mathbf{H} \mathbf{P} \mathbf{H}^T + \mathbf{R}^{}_{k}
\end{align}
   Here, $\mathbf{P}$ is the predicted covariance matrix, and $\mathbf{R}^{}_{k}$ is the mapped measurement noise covariance matrix.
\item Calculate Normalized Mahalanobis Distance:
\begin{align}
\label{eq_norml_maha}
D &= \mathbf{\epsilon}^T \mathbf{S}^{-1} \mathbf{\epsilon} + \ln \left| \mathbf{S} \right|
\end{align}
   Here, $\left| \mathbf{S} \right|$ represents the determinant of the matrix $\mathbf{S}$, and $\ln$ is the natural logarithm.
\end{enumerate}

As seen in Eq. \ref{eq_norml_maha}, we employed the normalized Mahalanobis distance, incorporating the logarithm of the determinant of the measurement covariance matrix. This ensures that data association decisions are not solely based on the discrepancies between measurements and predictions, but also holistically consider the accuracy and uncertainty of measurements. Consequently, this will yield more robust and reliable association decisions in object tracking.

\begin{table*}[t]
\centering

\begin{tabular}{l|p{0.85cm}p{0.85cm}p{0.85cm}p{0.85cm}|p{0.85cm}p{0.85cm}p{0.85cm}p{0.85cm}}
     \hline
    
     Tracker & \multicolumn{4}{c|}{\textbf{MOT17}} & \multicolumn{4}{c}{\textbf{MOT20}} \\
    \hline
    \textcolor{red}{appearance \& motion:} & HOTA$\uparrow$  & IDF1$\uparrow$  & MOTA$\uparrow$ & AssA$\uparrow$ & HOTA$\uparrow$  & IDF1$\uparrow$  & MOTA$\uparrow$ & AssA$\uparrow$\\
     \cline{2-5}
     \cline{5-9}
    FCG \cite{girbau2022multiple}& 62.6 & 77.7 & 76.7 & 63.4 & 57.3 & 69.7 & 68.0 & 58.1\\
   
    Quo Vadis \cite{dendorfer2022quo}& 63.1 & 77.7 & 80.3 & 62.1 & 61.5 & 75.7 & 77.8 & 59.9\\
    GHOST \cite{seidenschwarz2023simple} & 62.8 & 77.1 & 78.7	& - & 61.2 & 75.2 & 73.7	& -\\
    Bot-SORT \cite{aharon2022bot}& 65.0 & 80.2 & 80.5	& 65.5 & 63.3 & 77.5 & 77.8	& 62.9\\
    StrongSORT \cite{du2023strongsort}  & 64.4 & 79.5 & 79.6 & 64.4 & 62.6 & 77.0 & 73.8 & 64.0\\
    Deep OCSORT \cite{maggiolino2023deep} & 64.9 & 80.6 & 79.4 & 65.9 & \textbf{63.9} & \textbf{79.2} & 75.6 & \textbf{65.7}\\
    \hline
    \textcolor{red}{motion only:} &  HOTA$\uparrow$  & IDF1$\uparrow$  & MOTA$\uparrow$ & AssA$\uparrow$ & HOTA$\uparrow$  & IDF1$\uparrow$  & MOTA$\uparrow$ & AssA$\uparrow$\\
    \cline{2-5}
    \cline{5-9}
    ByteTrack \cite{zhang2022bytetrack} & 63.1 & 77.3 & 80.3 & 62.0 & 61.3 & 75.2 & 77.8 & 59.6\\
    C-BIoU \cite{yang2023hard}  & 64.1 & 79.7 & 81.1	& 63.7 & - & - & - & -\\  
    MotionTrack \cite{xiao2023motiontrack}  & 65.1 & 80.1 & \textbf{81.1}	& 65.1 & 62.8 & 76.5 & 78.0	& 61.8\\  
    SparseTrack \cite{liu2023sparsetrack}  & 65.1 & 80.1 & 81.0 & 65.1 & \textcolor{blue}{\textbf{63.4}} & 77.3 & \textbf{78.2} & 62.8\\  
    OCSORT  \cite{cao2022observation} & 63.2 & 77.5 & 78.0 & 63.4 & 62.4 & 76.3 & 75.7 & 62.5\\  
    \textbf{UCMCTrack (Ours)}& 64.3 & 79.0 & 79.0 & 64.6 & 62.8 & \textcolor{blue}{\textbf{77.4}} & 75.5 & \textcolor{blue}{\textbf{63.5}}\\
    \textbf{UCMCTrack+ (Ours)}& \textbf{65.8} & \textbf{81.1} & 80.5 & \textbf{66.6} & 62.8 & 77.4 & 75.7 & 63.4\\
    \hline 
\end{tabular}
\caption{Results on MOT17 \& MOT20 test. The detection results were obtained from ByteTrack \cite{zhang2022bytetrack}.}
\label{table1}
\end{table*}

\subsection{Process Noise Compensation}

In the context of MOT tasks, many previous works \cite{bewleySimpleOnlineRealtime2016,wojkeSimpleOnlineRealtime2017a,zhang2022bytetrack,cao2022observation} have treated the target's motion model as a Constant Velocity (CV) model without considering the noise impact caused by camera motion. However, camera motion is quite common in MOT tasks and can introduce significant noise that affects the tracking performance.
Assuming that the camera's motion-induced acceleration is the source of noise, we can represent it through the system motion model as follows:
\begin{align}
\label{eq12}
	\left\{\begin{array}{l}
		\Delta x =\frac{1}{2} \cdot \sigma \cdot (\Delta  t)^2\\
		\Delta v = \sigma \cdot \Delta t                                
	\end{array}\right.
\end{align}
where $\Delta x$ and $\Delta v$ represent the changes in position and velocity under the influence of noise, respectively. $\sigma$ denotes the acceleration change due to camera motion, and $\Delta t$ represents the time interval between two image frames.
Expressing Eq. \ref{eq12} in matrix form yields the matrix:
\begin{align}
\label{eq13}
        \mathbf{G}=
        \left[\begin{array}{ll}
		\frac{{\Delta t^2}}{2} & 0  \\
		\Delta t & 0  \\
		0 & \frac{{\Delta t^2}}{2} \\
            0 & \Delta t 
	\end{array}\right] 
\end{align}
It captures the relationship between the changes in position and velocity caused by noise in each direction.
For a two-dimensional CV Model with a Kalman filter, the covariance matrix of the process noise can be represented as follows: 
\begin{align}
\label{eq14}
\mathbf{Q_k} =  \mathbf{G}\cdot \text{diag}(\sigma_x, \sigma_y) \cdot \mathbf{G}^T 
\end{align}
where $\sigma_x$ and $\sigma_y$ denote the process compensation factors along the $x$ and $y$ axes, handling the motion noise cause by camera movements of tilt and rotation respectively.

\section{Experiments}

\subsection{Setting}
\subsubsection{Datasets}
We conducted a fair evaluation of UCMCTrack on multiple publicly available datasets, including MOT17  \cite{milan2016mot16}, MOT20 \cite{dendorfer2020mot20}, DanceTrack \cite{Sun_2022_CVPR}, and KITTI \cite{geiger2013vision}. 
Both MOT17 and MOT20 are pedestrian tracking datasets, and their motion is mostly linear. It is worth noting that MOT20 has a significantly higher density of pedestrians, making it a challenging dataset for tracking. The primary task of the DanceTrack \cite{Sun_2022_CVPR} is to track dancers, who not only have similar appearances but also perform a large number of irregular movements.The KITTI  \cite{geiger2013vision} is an autonomous driving dataset, and we only utilized the left color camera images for the visual vehicle and pedestrian tracking task. Compared to other datasets, KITTI has a lower frame rate, only 10 FPS, and the camera's motion is more intense.

\begin{table}[t]
\centering
\begin{tabular}{l|p{0.75cm}p{0.75cm}p{0.75cm}p{0.75cm}}
     \hline
     Tracker   & HOTA$\uparrow$  & IDF1$\uparrow$  & MOTA$\uparrow$ & AssA$\uparrow$\\
    \hline
    \textcolor{red}{appearance \& motion:}& \\
    FCG   & 48.7 & 46.5 & 89.9 & 29.9 \\    
    GHOST   & 56.7  & 57.7 & 91.3	& 39.8 \\
    StrongSORT     & 55.6 & 55.2 & 91.1 & 38.6 \\
    Deep OCSORT   & 61.3 & 61.5 & \textbf{92.3} & 45.8 \\
    \hline
    \textcolor{red}{motion only:} & \\
    ByteTrack    & 47.3 & 52.5 & 89.5 & 31.4 \\
    C-BIoU    & 60.6 & 61.6 & 91.6	& 45.4 \\  
    MotionTrack  & 58.2 & 58.6 & 91.3 & 41.7 \\  
    SparseTrack   & 55.5 & 58.3 & 91.3 & 39.1 \\  
    OCSORT    & 55.1 & 54.9 & \textcolor{blue}{\textbf{92.2}} & 40.4 \\  
    \textbf{UCMCTrack  (Ours)} &  63.4 & \textbf{65.0} & 88.8 & 51.1 \\
    \textbf{UCMCTrack+ (Ours)} & \textbf{63.6} & \textbf{65.0} & 88.9 & \textbf{51.3} \\
    \hline 
\end{tabular}
\caption{Results on DanceTrack-test. The detection results were obtained from ByteTrack \cite{zhang2022bytetrack}.}
\label{table2}
\end{table}

\begin{table*}[t]
\centering
\begin{tabular}{l|ccc|ccc}
     \hline
     Tracker & \multicolumn{3}{c|}{\textbf{Car}} &\multicolumn{3}{c}{\textbf{Pedestrain}}\\
    \hline
    \textcolor{red}{appearance \& motion:} & HOTA$\uparrow$  & MOTA$\uparrow$ & AssA$\uparrow$ & HOTA$\uparrow$  & MOTA$\uparrow$ & AssA$\uparrow$ \\

    QD-3DT \cite{hu2022monocular} &72.8 & 85.9 & 72.2 & 41.1 & 51.8 &38.8 \\    
    TuSimple \cite{choi2015near,He_2016_CVPR} &71.6 & 86.3 & 71.1 & 45.9 & 57.6 &47.6\\
    StrongSORT \cite{du2023strongsort}   &\textbf{77.8} & 90.4 & \textbf{78.2} & 54.5 & 67.4 &57.3 \\
    \hline
    \textcolor{red}{motion only:} & HOTA$\uparrow$  & MOTA$\uparrow$ & AssA$\uparrow$ & HOTA$\uparrow$  & MOTA$\uparrow$ & AssA$\uparrow$ \\
    CenterTrack \cite{zhou2020tracking}  &73.0 & 88.8 & 71.2 & 40.4 & 53.8 &36.9 \\
    TrackMPNN \cite{rangesh2021trackmpnn}   &72.3 & 87.3 & 70.6 & 39.4	& 52.1 & 35.5 \\  
    OCSORT  \cite{cao2022observation}  &76.5 & 90.3 & 76.4 & 54.7 & 65.1 & \textbf{59.1} \\ 
    \textbf{UCMCTrack (Ours)} & \textcolor{blue}{\textbf{77.1}}  &\textbf{90.4} & \textcolor{blue}{\textbf{77.2}} & \textbf{55.2} & \textbf{67.4} & 58.0 \\
   \textbf{UCMCTrack+  (Ours)} & 74.2  &90.2 & 71.7 & 54.3 & 67.2 & 56.3 \\
    \hline 
\end{tabular}
\caption{Results on KITTI-test.The detection results were obtained from PermaTrack \cite{Tokmakov_2021_ICCV}.}
\label{table3}
\end{table*}

\subsubsection{Metrics}
 We employ the CLEAR metrics  \cite{bernardin2008evaluating} which include MOTA, FP, FN, and others, along with IDF1 \cite{ristani2016performance} and TA \cite{luiten2021hota}, to evaluate the tracking performance comprehensively in various aspects.
MOTA emphasizes the detector's performance, while IDF1 measures the tracker's ability to maintain consistent IDs. We also emphasize the use of AssA to evaluate the association performance. On the other hand, HOTA achieves a balance between detection accuracy, association accuracy, and localization accuracy, making it an increasingly important metric for evaluating trackers.

\subsubsection{Implementation Details}
For fair comparison, we directly used the existing baseline object detection method YOLOX \cite{DBLP:journals/corr/abs-2107-08430}. The weight files for MOT17, MOT20 and DanceTrack were obtained from ByteTrack \cite{zhang2022bytetrack}. For KITTI, we used the detection results from PermaTrack \cite{Tokmakov_2021_ICCV}. We applied the Enhanced Correlation Coefficient maximization (ECC) \cite{4515873} model for camera motion compensation, which is consistent with strongSORT \cite{du2023strongsort}. For MOT17, MOT20, and DanceTrack datasets, we manually estimated the camera parameters since they are not publicly accessible. In contrast, KITTI dataset readily furnishes the requisite camera parameters.

\subsection{Benchmark Evaluation}
Here, we present the benchmark results for multiple datasets. $\uparrow$/$\downarrow$ indicate that higher/lower is better, respectively. The highest scores for each group are highlighted in \textbf{bold}, and the highest score for the motion group is marked in \textcolor{blue}{\textbf{blue}}. “UCMCTrack+” denotes the enhancement of UCMCTrack with the
additional incorporation of CMC.

\subsubsection{MOT17 and MOT20}
Our UCMCTrack results on MOT17 and MOT20 datasets are presented in Table \ref{table1}, respectively. We used a private detector to generate the detection results and ensured fairness by aligning the detections with OC-SORT \cite{cao2022observation} and ByteTrack \cite{zhang2022bytetrack}. UCMCTrack+ has attained state-of-the-art (SOTA) performance, notably on the MOT17 dataset, surpassing the SOTA methods by 0.9 in HOTA, 0.5 in IDF1, and 0.7 in AssA. It even surpasses leading algorithms that leverage both motion and appearance features at a considerable margin, highlighting its effective use of motion information to enhance the robustness and efficiency of the tracking.

\subsubsection{DanceTrack}
To demonstrate UCMCTrack's performance under irregular motion scenarios, we present the test set results on DanceTrack in Table \ref{table2}. UCMCTrack+ outperforms the SOTA methods with an improvement of 2.3 in HOTA, 3.4 in IDF1, and 5.5 in AssA. This highlights the effectiveness of our tracker in handling targets with irregular motions and further validates its SOTA performance.

\subsubsection{KITTI}
In Table \ref{table3}, we present the results of UCMCTrack on the KITTI dataset. It's noteworthy that the addition of CMC on the KITTI dataset did not yield favorable results. We believe that this might be due to the inaccuracies present in the CMC parameters. This observation indicates that UCMC demonstrates stronger generalization capabilities than CMC, particularly in complex scenarios.The performance of UCMCTrack in the KITTI dataset demonstrates its effectiveness in addressing challenges posed by high-speed motion and low frame-rate detections.

\subsection{Ablation Studies on UCMC}
The ablation experiments of UCMCTrack were conducted on the validation sets of MOT17 and DanceTrack. For MOT17, the validation set was split following the prevailing conventions \cite{zhou2020tracking}. The baseline is chosen as ByteTrack \cite{zhang2022bytetrack}.

\subsubsection{Component Ablation}
We  undertook a comprehensive validation of UCMCTrack's key components: Mapped Mahalanobis Distance (MMD), Correlated Measurement Distribution (CMD), and Process Noise Compensation (PNC), testing on both MOT17 and DanceTrack datasets. As demonstrated in Table \ref{table4}, each component of UCMCTrack plays a crucial role in enhancing its overall efficacy.
A noteworthy observation is the decline in performance when transitioning from IoU to MMD. This can be attributed to the IoU's consideration of both position and height information, whereas MMD solely focuses on position, leading to the observed decline in outcomes. However, when CMD is applied post MMD, it better utilizes the distribution information, thus surpassing the performance of the IoU-based baseline. This observation also underscores the potential benefits of integrating height information into MMD for future research.
Lastly, the incorporation of PNC effectively mitigates the motion noise introduced by camera movements, elevating the tracking performance to state-of-the-art levels.

\begin{table}[ht]
\centering

\begin{tabular}{lp{0.2cm}p{0.6cm}p{0.45cm}p{0.4cm}p{0.6cm}p{0.8cm}}
    \hline 
    Method & IoU & MMD & CMD & PNC   & IDF1$\uparrow$ & HOTA$\uparrow$  \\
    \hline 
    \multicolumn{7}{c}{\textbf{MOT17 Validation Set}}\\   
    \hline
    baseline & \checkmark & - & - & - & 77.10 & 68.43 \\
    UCMCTrack-v1 & - & \checkmark & - & - & 75.88&68.09 \\
    UCMCTrack-v2 & - & \checkmark & \checkmark & - & 79.68&70.44 \\
    UCMCTrack-v3 & - & \checkmark & \checkmark & \checkmark & \textbf{82.20}&\textbf{71.96}\\
    \hline
    \multicolumn{7}{c}{\textbf{DanceTrack Validation Set}}\\ 
    \hline
    baseline & \checkmark & - & - & - & 47.27&47.93	 \\
    UCMCTrack-v1 & - & \checkmark & - & - & 43.76&46.32\\
    UCMCTrack-v2 & - & \checkmark & \checkmark & - & 53.93&55.06	 \\
    UCMCTrack-v3 & - & \checkmark & \checkmark & \checkmark & \textbf{62.64}&\textbf{60.42}	\\
    \hline
\end{tabular}
\caption{Ablation of UCMCTrack components.}
\label{table4}
\end{table}

\begin{table}[ht]
\centering

\begin{tabular}{lcccc}
    \hline 
    Method & CMC & UCMC   & IDF1$\uparrow$ & HOTA$\uparrow$ \\
    \hline 
    \multicolumn{5}{c}{\textbf{MOT17 Validation Set}}\\   
    \hline
    baseline & - & - & 77.10 & 68.43 \\
    baseline+CMC & \checkmark & - & 81.07 & 70.97 \\
    UCMCTrack & - & \checkmark & 82.20 & 71.96 \\
    UCMCTrack+ & \checkmark & \checkmark & \textbf{84.05} & \textbf{72.97}\\
    \hline
    \multicolumn{5}{c}{\textbf{DanceTrack Validation Set}}\\ 
    \hline
    baseline & - & - & 47.27 & 47.93 \\
    baseline+CMC & \checkmark & - & 46.74 & 47.55 \\
    UCMCTrack & - & \checkmark & \textbf{62.64} & \textbf{60.42} \\
    UCMCTrack+ & \checkmark & \checkmark & 62.52 & 59.18\\
    \hline
    
\end{tabular}
\caption{CMC ablation.}
\label{table5}
\end{table}

\begin{figure*}[htbp]
  \centering
  \begin{subfigure}[b]{0.33\textwidth}
    \centering
    \includegraphics[width=\textwidth]{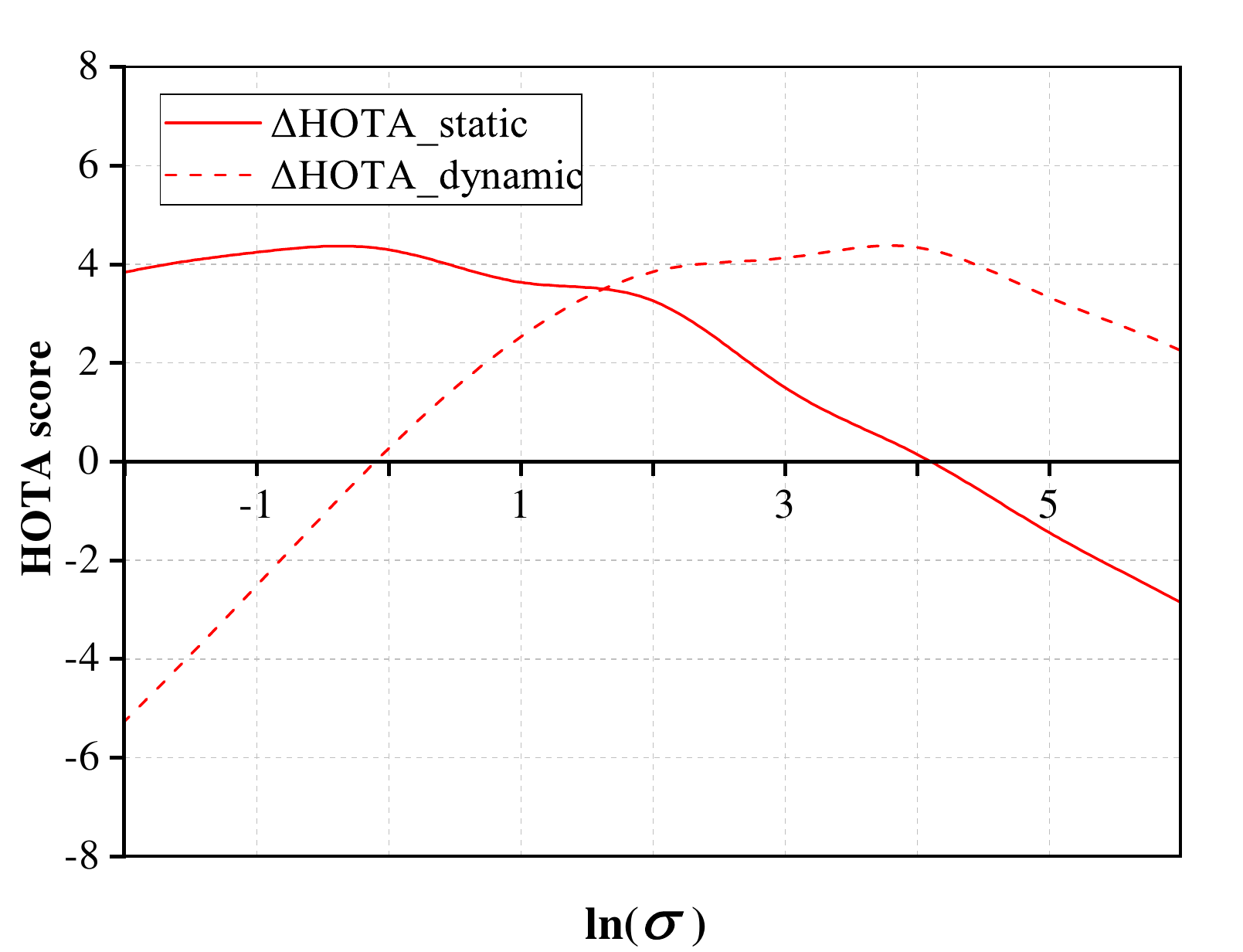}
    \caption{Process Compensation Factor}
    \label{fig:subfig1}
  \end{subfigure}
  \begin{subfigure}[b]{0.33\textwidth}
    \centering
    \includegraphics[width=\textwidth]{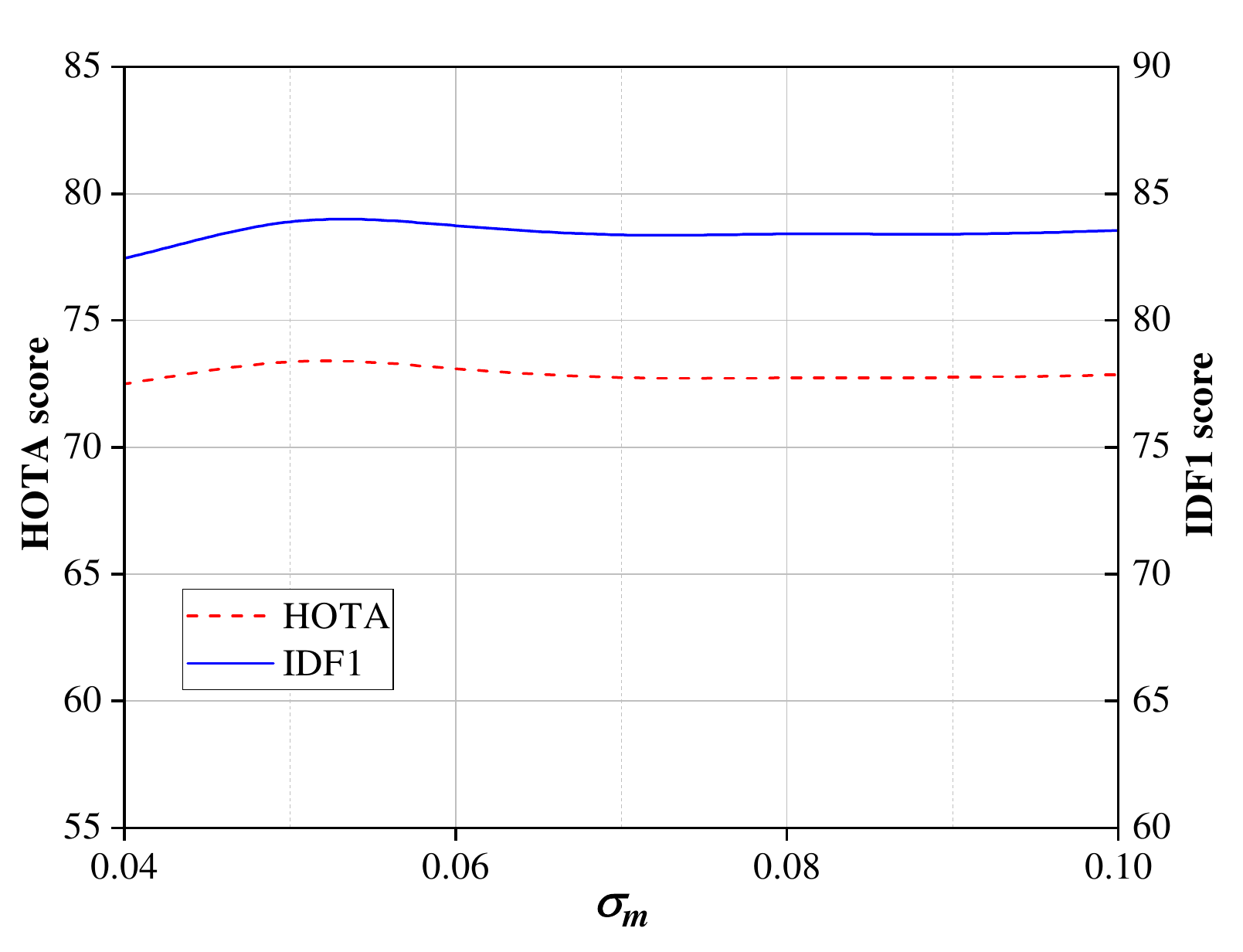}
    \caption{Detector Noise Factor}
    \label{fig:subfig2}
  \end{subfigure}
   \begin{subfigure}[b]{0.33\textwidth}
    \centering
    \includegraphics[width=\textwidth]{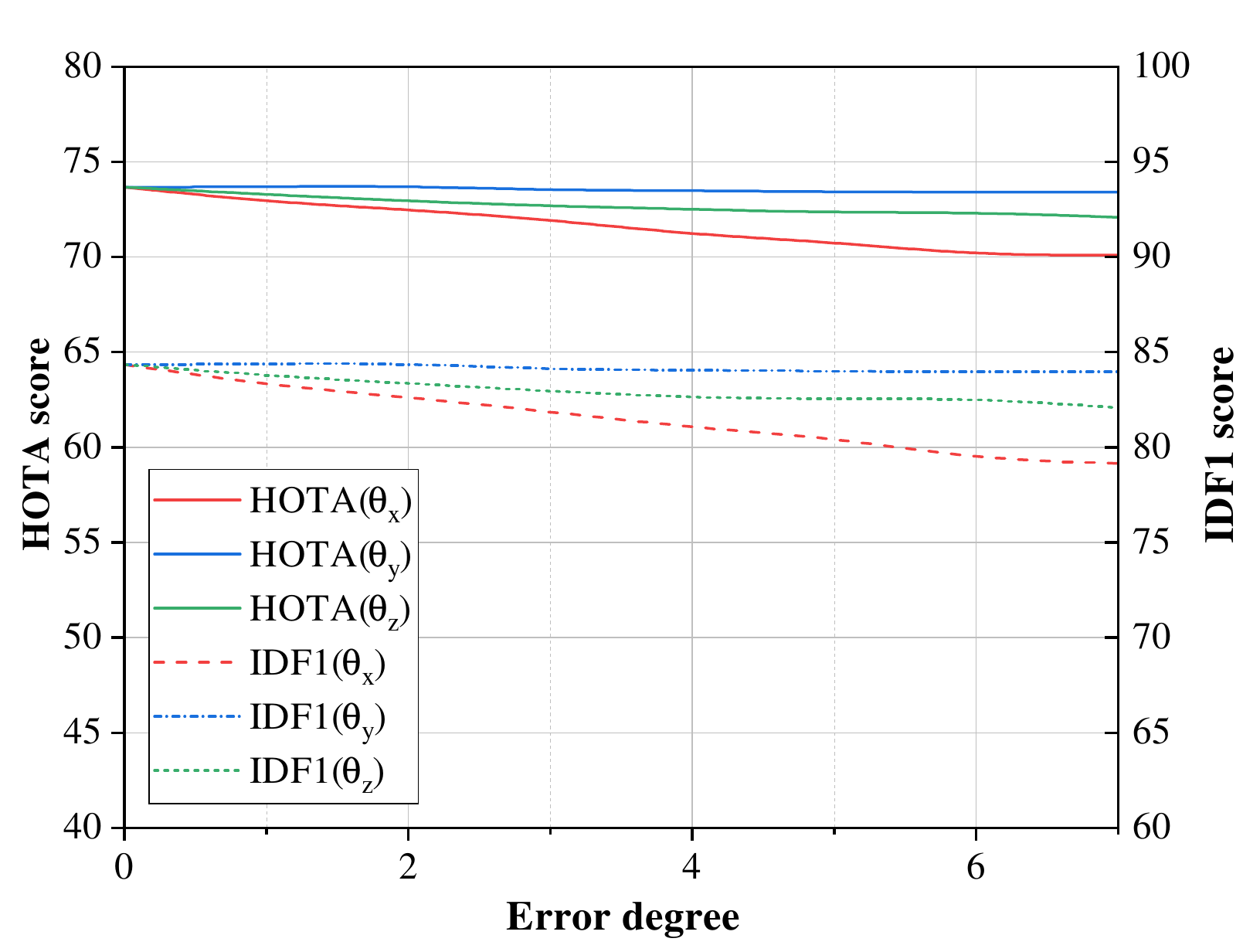}
    \caption{Camera Parameters Error}
    \label{fig:subfig3}
  \end{subfigure}
  \caption{In-depth analysis of key parameters and robustness in UCMCTrack.}
  \label{fig:twosubfigures}
\end{figure*}

\subsubsection{CMC Ablation}
To explore the impact of CMC on UCMCTrack, we conducted ablation experiments on MOT17 and DanceTrack validation datasets, as shown in Table \ref{table5}. In MOT17, employing UCMC results in a greater performance enhancement compared to using the baseline combined with CMC, underscoring the effective role of UCMC in compensating for camera motion. Furthermore, a subsequent application of CMC to UCMC yields an additional performance boost, consistent with the results observed on the MOT17 test set.
Interestingly, upon employing CMC on DanceTrack, the performance of the baseline and UCMC algorithm actually deteriorated. This can be attributed to the fact that most video sequences on DanceTrack don't exhibit pronounced viewpoint shifts. After employing CMC, minor detection offsets for overlapping targets might result in matching with an incorrect trajectory, culminating in a mild performance setback. In contrast, the use of UCMC resulted in a significant performance boost, suggesting its effectiveness in scenes with only minor camera jitters.

\subsubsection{Influence of Process Compensation Factor}
In order to further investigate the impact of the process compensation factors on the performance of UCMCTrack, we divided the MOT17 validation sequences into dynamic and static scenes. The horizontal axis represents the natural logarithm of the compensation factor, with both $\sigma_x$ and $\sigma_y$ set to $\sigma$ as per Eq. \ref{eq14}, and the vertical axis represents the difference in HOTA compared to the baseline. 
As illustrated in Figure \ref{fig:subfig1}, the influence of the compensation factor on dynamic and static scenes exhibits two distinct patterns. This divergence underscores the necessity of tailoring the compensation parameters separately for each scene type. Adapting these parameters specific to the scene's nature ensures that the tracker operates at its optimum performance. This distinction also highlights the role of the compensation factor in mitigating the effects of camera movements. For static scenes, a smaller compensation factor is recommended, while for dynamic scenes, a larger compensation factor is essential to counteract the impact of camera motion, thus enhancing the tracking performance.

\subsubsection{Influence of Detector Noise Factor}
We conducted ablation studies on the hyperparameter $\sigma_{m}$ in Eq. \ref{eq2} to explore its impact on UCMCTrack. As shown in Figure \ref{fig:subfig2}, the tracker's performance varies with different $\sigma_{m}$. When the $\sigma_{m}$ is set to 0.05, the HOTA and IDF1 metrics reach their highest values. It is evident that the influence of the $\sigma_{m}$ on HOTA and IDF1 remains relatively minor within the range of 0.04 to 0.1. This indicates that our UCMCTrack is not sensitive to $\sigma_{m}$.

\subsubsection{Robustness to Camera Parameters Error}
Our method, UCMCTrack, relies on the camera parameters to project targets from the image plane to the ground plane. However, in scenarios where camera parameters are not provided, we manually estimate them. This means that the estimated camera parameters may not be accurate. We conducted separate ablation experiments for the camera extrinsics errors along the X, Y, and Z axes, and the results are shown in Figure \ref{fig:subfig3}.
We observed that errors in the Y axis have a minor impact on performance. On the other hand, errors in the X and Z axes have more significant effects. This is due to the Y-axis corresponding to the yaw direction, variations in this direction have a less impact on the deformation of the estimated ground plane. However, When there are substantial errors in the estimated camera extrinsics along the X and Y axes, the performance of UCMCTrack notably degrades. This can be attributed to significant deformations in the estimated ground plane. Under such circumstances, adjustments to the camera parameters are required to ensure that the estimated ground plane aligns closely with the actual terrain.

\section{Conclusion}
In this work, we presented UCMCTrack, a state-of-the-art multi-object tracker demonstrating superior performance across a variety of datasets. UCMCTrack employs a novel distance measure based on normalized mahalanobis distance on mapped ground plane, marking a significant departure from the conventional reliance on IoU. This innovation enables the tracker to adeptly handle the challenges introduced by camera movements, using only a consistent set of compensation parameters across a single video sequence. 
However, an inherent limitation of UCMCTrack is its assumption that targets reside on a singular ground plane. Looking forward, there is considerable potential to enhance UCMCTrack's effectiveness by integrating it with well-established distance measures such as IoU and ReID. We believe that this will lay the groundwork for subsequent advancements in the future research of motion-based multi-object tracking.

\section{Acknowledgments}

This work was supported by the National Key Research and Development Program of China under Grant 2022YFC3803700, in part by the Natural Science Foundation of China under Grants 52002036, the Changsha Science and Technology Major Project under Grant kh2202002 and kh2301004, the Hunan Provincial Natural Science Foundation of China under Grant 2022JJ30611, the Scientific Research Fund of Hunan Provincial Education Department under Grant 21B0342.

\bibliography{aaai24}


\clearpage
\appendix

\section{Appendix A}
\subsection{Pseudo-code of UCMCTrack}
\begin{algorithm}[ht]
\caption{Pseudo-code of UCMC}
\label{alg:algorithm}
\textbf{Input}: A video sequence: $V$, Object detector result: $Dets$, Camera parameter: $Para$\\
\textbf{Parameter}: Confidence threshold:$\tau$, Lost time threshold:$dt$, The process compensation factors:$\sigma_x,\sigma_y$, Detection noise factor:$\sigma_m$\\
\textbf{Output}: Tracks $\mathcal{T}$  of the sequence
\begin{algorithmic}[1] 
\WHILE{ frame $f_k$ not None}
 \STATE $\mathcal{T} \gets $ KF($\mathcal{T},\sigma_x,\sigma_y$) 
 \COMMENT{Updating the trajectory positions and distribution}
\STATE Let $\mathcal{C} \gets \emptyset $.\COMMENT {Initializing the cost matrix}
\STATE $\mathcal{D} \kappa \gets Dets(f_k)$ \COMMENT{Get detection result}
\FOR{$\mathcal{D}_i$ in $\mathcal{D} \kappa$}
 \STATE $\mathcal{Y}_i , \mathcal{R}_i \gets$  Map($\mathcal{D}_i$,$Para,\sigma_m$) \COMMENT{Get the position and distribution of $\mathcal{D} \kappa$}
 \FOR{$\mathcal{T}_j$ in $\mathcal{T}$}
   \STATE $\mathcal{C}_{ij} \gets$ MahaD($\mathcal{Y}_i , \mathcal{R}_i, \mathcal{T}_j$) 
 \ENDFOR
\ENDFOR
\STATE  $\mathcal{T}_{Associate}, \mathcal{T}_{remain}, \mathcal{Y}_{remain} \gets  $ Hungarian($\mathcal{C}$)
\FOR{ $t$ in $\mathcal{T}_{remain}$ }
   \IF { $t$ miss time $ > dt$ }
   \STATE delete $t$
   \ELSE
   \STATE $t \gets$  $t$ miss time +1 
   \ENDIF
\ENDFOR
\FOR{ $y$ in $\mathcal{Y}_{remain}$ }
   \IF { The confidence level of $y$  $ < \tau$ }
   \STATE delete $y$
   \ELSE
   \STATE Initializing a new trajectory, $t$
   \ENDIF
\ENDFOR
\ENDWHILE
\STATE \textbf{return} Tracks $\mathcal{T}$
\end{algorithmic}
\end{algorithm}

We introduce UCMCTrack, a simple, efficient, and powerful tracker that achieves state-of-the-art performance without IoU. In contrast to existing methods that often require frame-by-frame camera motion compensation to mitigate issues caused by camera rotations or shakes, our approach utilizes a compact set of parameters to achieve remarkable results with enhanced generalization and practicality. The pseudocode for UCMCTrack is outlined in Algorithm \ref{alg:algorithm}:

The input of UCMCTrack is a video sequence $V$, along with an object detector $Det$ and camera parameter $Para$. We also set a detection score threshold $\tau$,lost time threshold $dt$,the process compensation factors $\sigma_x,\sigma_y$, and detection noise factor $\sigma_m$. The output of UCMCTrack is the tracks $\mathcal{T}$ of the video and each track contains the bounding box and identity of the object in each frame.

For each input frame, we start by predicting the previously retained trajectories using a constant velocity motion model. This prediction provides us with the anticipated trajectory positions and their distribution on the ground plane.(line 2 in Algorithm \ref{alg:algorithm})

We compute the distances between the detection boxes and the trajectories, forming the association cost matrix $\mathcal{C}$. Initially, we map the detection results $\mathcal{D} \kappa$ to the ground plane using camera parameters, obtaining their positions and distributions. Then, we calculate the normalized Mahalanobis distance between the detection and all predicted trajectories. Finally, these results are stored in the association cost matrix $\mathcal{C}$.(line 3 to 10 in Algorithm \ref{alg:algorithm})

The Hungarian algorithm is utilized for optimal assignment. By using the association cost matrix $\mathcal{C}$ as input, the Hungarian algorithm is employed to perform assignment. This process yields associated trajectories, unassociated tracklets, and unassociated detection boxes. The associated trajectories are retained for preparation in the subsequent frames of tracking. Unassociated tracklets experience an increase of one in their "lost time". When the lost time surpasses the threshold $dt$, the tracklets will be deleted; otherwise, the tracklets will be categorized into the associated trajectories. Regarding unassociated detection boxes, if their confidence score exceeds the confidence threshold $\tau$, they enter the tracklet initialization stage.(line 11 to 25 in Algorithm \ref{alg:algorithm})

\section{Appendix B}
\subsection{Mapped Coordinates and Distribution}
By idealizing the camera model as a linear camera model \cite{yu2004general}, we can derive an expression that relates three-dimensional world coordinates to two-dimensional pixel coordinates based on the geometric relationship of the linear camera model.

\begin{align}
   \
	\label{ap:eq1}
       {\gamma}_0
	\left[\begin{array}{l}
		u \\
		v \\
		1
	\end{array}\right]
        =\left[\begin{array}{cccc}
        f_{x} & 0 & u_{0} & 0 \\
        0 & f_{y} & v_{0} & 0 \\
        0 & 0 & 1 & 0
        \end{array}\right]\left[\begin{array}{cc}
        \mathbf{R} & \mathbf{T} \\
        0 & 1
        \end{array}\right]\left[\begin{array}{c}
        x_{} \\
        y_{} \\
        z_{} \\
        1
        \end{array}\right]
\end{align}
Where $x$, $y$, and $z$ represent three-dimensional coordinates, $u$ and $v$ represent pixel coordinates in the image, and ${\gamma}_0$ is the scale factor.

Typically, the two matrices in equation \ref{ap:eq1} are defined as the intrinsic matrix $\mathbf{K}_i$ and the extrinsic matrix $\mathbf{K}_o$, as illustrated in equation \ref{ki} and \ref{ko}:

\begin{align}
   \
	\label{ki}
        &\mathbf{K}_i=\left[\begin{array}{cccc}
        f_{x} & 0 & u_{0} & 0 \\
        0 & f_{y} & v_{0} & 0 \\
        0 & 0 & 1 & 0
        \end{array}\right]\\
        \label{ko}
        &\mathbf{K}_o=\left[\begin{array}{cc}
        \mathbf{R} & \mathbf{T} \\
        0 & 1
        \end{array}\right]
\end{align}
Where $\mathbf{K}_o$ incorporates the camera extrinsic errors mentioned in the text for the ablation studies on the x, y, and z axes.

By substituting equations \ref{ki} and \ref{ko} into equation \ref{ap:eq1}, we can obtain:

\begin{align}
   \
	\label{ap:eq15}
       {\gamma}_0
	\left[\begin{array}{l}
		u \\
		v \\
		1
	\end{array}\right]
       ={\mathbf{K}_i}{\mathbf{K}_o}
       \left[\begin{array}{l}
		x_{} \\
		y_{} \\
           z_{} \\
		1
	\end{array}\right] 
\end{align}

If we compute the product of $\mathbf{K}_i$ and $\mathbf{K}_o$, equation \ref{ap:eq15} can be represented as:
\begin{align}
   \
	\label{ap:eq16}
       {\gamma}_0
	\left[\begin{array}{l}
		u \\
		v \\
		1
	\end{array}\right]
       =\left[\begin{array}{llll}
		{\theta}_{11} & {\theta}_{12} & {\theta}_{13} & {\theta}_{14} \\
		{\theta}_{21} & {\theta}_{22} & {\theta}_{23} & {\theta}_{24} \\
		{\theta}_{31} & {\theta}_{32} & {\theta}_{33} & {\theta}_{34} 
	\end{array}\right]
       \left[\begin{array}{l}
		x_{} \\
		y_{} \\
           z_{} \\
		1
	\end{array}\right]
\end{align}

Let $z = z_0$ be a constant. Rewriting equation \ref{ap:eq16}:
\begin{align}
\label{ap:eq13}
       \left[\begin{array}{l}
		u \\
		v \\
		1
	\end{array}\right]
       &=\left[\begin{array}{lll}
		{\theta}_{11} & {\theta}_{12}  & {\theta}_{13}z_0+{\theta}_{14} \\
		{\theta}_{21} & {\theta}_{22}  & {\theta}_{23}z_0+{\theta}_{24} \\
		{\theta}_{31} & {\theta}_{32}  & {\theta}_{33}z_0+{\theta}_{34} 
	\end{array}\right] \frac{1}{\gamma}
       \left[\begin{array}{c}
		x_{} \\
		y_{} \\
		1
	\end{array}\right] \notag\\
       &=\mathbf{A}\frac{1}{\gamma}
       \left[\begin{array}{c}
		x_{} \\
		y_{} \\
		1
	\end{array}\right]
\end{align}
Where, $\mathbf{A}$ is denoted as:
\begin{align}
\mathbf{A} = 
\left[\begin{array}{lll}
		{\theta}_{11} & {\theta}_{12}  & {\theta}_{13}z_0+{\theta}_{14} \\
		{\theta}_{21} & {\theta}_{22}  & {\theta}_{23}z_0+{\theta}_{24} \\
		{\theta}_{31} & {\theta}_{32}  & {\theta}_{33}z_0+{\theta}_{34} 
	\end{array}\right]
\end{align}
Let:
\begin{align}
	\label{ap:eq3}
       b= \mathbf{A}^{-1}
	\left[\begin{array}{l}
		u \\
		v \\
		1
	\end{array}\right]=
       \left[\begin{array}{c}
		b_{1} \\
		b_{2} \\
		b_{3}
	\end{array}\right]
\end{align}
Thus, the physical coordinates can be computed from pixel coordinates as follows:

\begin{align}
	\label{ap:eq4}
	\left[\begin{array}{l}
		x_{} \\
		y_{} \\
	\end{array}\right]=
       \left[\begin{array}{c}
		\frac {b_{1}}{b_{3}} \\
		\frac {b_{2}}{b_{3}} \\
	\end{array}\right]
\end{align}

Once we obtain the computation formula from pixel coordinates to ground coordinates, the errors in pixel coordinates will also be mapped to ground coordinates. Therefore, it becomes necessary to further explore the distribution of detection errors in the ground plane caused by the errors in the image plane.
Let's express the inverse matrix of $A$ in equation \ref{ap:eq5}:

\begin{align}
\label{ap:eq5}
       \mathbf{A}^{-1}=
       \left[\begin{array}{lll}
		{a}_{11} & {a}_{12} & {a}_{13}  \\
		{a}_{21} & {a}_{22} & {a}_{33}  \\
		{a}_{31} & {a}_{32} & {a}_{33}  
	\end{array}\right] 
\end{align}

By combining equations \ref{ap:eq3}, \ref{ap:eq4}, \ref{ap:eq5}, we can obtain expressions for $x_w$ and $y_w$:

\begin{align}
\label{ap:eq6}
	\left\{\begin{array}{l}
		x_{}={\gamma}\left(a_{11} u+a_{12} v+a_{13}\right) \\
		y_{}={\gamma}\left(a_{21} u+a_{22} v+a_{23}\right)
	\end{array}\right.
\end{align}
Where $\gamma$ is defined as the reciprocal of $b_3$.

Taking the total derivative of equation \ref{ap:eq6}:
\begin{align}
\label{ap:eq7}
	\left\{\begin{array}{l}
		d x_{}={\gamma}\left(a_{11}d u+a_{12}d v\right) +b_1 d \gamma\\
		d y_{}={\gamma}\left(a_{21}d u+a_{22}d v\right) +b_2 d \gamma\\
        d \gamma=- \left( \frac{1}{{b_3}} \right)^2 \left( a_{31} \, du + a_{32} \, dv \right)
	\end{array}\right.
\end{align}

Combined with equations \ref{ap:eq4}, \ref{ap:eq7}, we have
\begin{align}
\left.\left[\begin{matrix}dx\\dy\end{matrix}\right.\right]
=
\left[ {\begin{array}{*{20}{c}}
    {\gamma {a_{11}} - {a_{31}}{\gamma }{x}}&{\gamma {a_{12}} - {a_{32}}{\gamma}{x}}\\
    {\gamma {a_{21}} - {a_{31}}{\gamma }{y}}&{\gamma {a_{22}} - {a_{32}}{\gamma }{y}}
    \end{array}} \right]\left[\begin{matrix}du\\dv\end{matrix}\right]
\end{align}

Let's assume that the measurement errors of detectors on the image plane follow an independent normal distribution, and their covariance matrix $R^{uv} _{k}$ can be represented as:
\begin{align}
\label{ap:eq2}
    \mathbf{R}^{uv} _{k}  = \left[ {\begin{array}{*{20}{c}}
    {(\sigma_{m} {w}_{k})^2}&{0}\\
    {0}&{(\sigma_{m} {h}_{k})^2}\end{array}
    } \right]
\end{align}
Where: $\sigma_{m}$ represents the detection noise factor as a hyperparameter, and ${w}_{k}$ and ${h}_{k}$ denote the detected width and height from the detector.

Then, the covariance matrix of measurement errors in the ground plane would be:
\begin{align}
\mathbf{R}^{}_{k}= \mathbf{C}{\mathbf{R}^{uv}_{k} }{\mathbf{C}^T}
\end{align}
where:
\begin{align}
    \mathbf{C} 
    =\left[ {\begin{array}{*{20}{c}}
    {\gamma {a_{11}} - {a_{31}}{\gamma }{x}}&{\gamma {a_{12}} - {a_{32}}{\gamma}{x}}\\
    {\gamma {a_{21}} - {a_{31}}{\gamma }{y}}&{\gamma {a_{22}} - {a_{32}}{\gamma }{y}}
    \end{array}} \right]
\end{align}
Thus, we obtain the mapped measurement noise matrix $\mathbf{R}_k$ in the ground plane.

\section{Appendix C}
\subsection{Kalman Filter Model}
In the domain of object tracking, where no active control is present, the discrete-time Kalman filter is guided by the following set of linear stochastic difference equations:
\begin{align}
\mathbf{x}_k &= \mathbf{F}_k \mathbf{x}_{k-1}+\mathbf{n}_{k-1}\\
\mathbf{z}_k &= \mathbf{H}_k \mathbf{x}_k+\mathbf{v}_{k}
\end{align}
Where $\mathbf{F}_k$ is the transition matrix from discrete-time
$k-1$ to $k$. The observation matrix is $H_k$. The random
variables $\mathbf{n}_{k}$ and $\mathbf{v}_{k}$ represent the process and measurement
noise respectively. They are assumed to be independent and
identically distributed (i.i.d) with normal distribution.
\begin{align}
\mathbf{n}_{k} \sim \mathcal{N}\left(\mathbf{0}, \mathbf{Q}_{k}\right), \quad \mathbf{v}_{k} \sim \mathcal{N}\left(\mathbf{0}, \mathbf{R}_{k}\right)
\end{align}

The Kalman filter consists of prediction and update steps. The entire Kalman filter can be summarized by the following recursive equations:
\begin{enumerate}
\item prediction:
\begin{align}
\label{KF_pre:1}
\hat{\mathbf{x}}_{k \mid k-1}&=\mathbf{F}_{k} \hat{\mathbf{x}}_{k-1 \mid k-1} \\
\mathbf{P}_{k \mid k-1}&=\mathbf{F}_{k} \mathbf{P}_{k-1 \mid k-1} \mathbf{F}_{k}^{\top}+\mathbf{Q}_{k}
\end{align}

\item update:
\begin{align}
&\mathbf{K}_{k} = \mathbf{P}_{k \mid k-1}\mathbf{H}_{k}^{\top}(\mathbf{H}_{k}\mathbf{P}_{k \mid k-1}\mathbf{H}_{k}^{\top}+\mathbf{R}_{k})^{-1}\\
&\hat{\mathbf{x}}_{k \mid k}=\hat{\mathbf{x}}_{k \mid k-1}+\mathbf{K}_{k}(\mathbf{z}_k-\mathbf{H}_{k}\hat{\mathbf{x}}_{k \mid k-1})\\
&\mathbf{P}_{k \mid k}=(\mathbf{I}-\mathbf{K}_{k}\mathbf{H}_{k})\mathbf{P}_{k \mid k-1}
\end{align}

\end{enumerate}
At each step $k$, KF predicts the prior estimate of state $\hat{\mathbf{x}}_{k \mid k-1}$ and the covariance matrix $\mathbf{P}_{k \mid k-1}$. KF updates the posterior state estimation $\hat{\mathbf{x}}_{k \mid k}$ given the observation $\mathbf{z}_k$ and the estimated covariance $\mathbf{P}_{k \mid k}$ , calculated based on the optimal Kalman gain $\mathbf{K}_{k}$.

In UCMCTrack, the observation is denoted as $\mathbf{z} = [x, y]^\top$ and the state vector is $\mathbf{x} = [x, \dot{x}, y, \dot{y}]^\top$. The $\mathbf{Q}_{k}$ is provided with a detailed expression in equation 11 of the main text, and $\mathbf{R}_{k}$ is referenced from equation \ref{ap:eq4}.The corresponding transition matrix and observation matrix for the constant velocity model are given by Equation \ref{ap:eq19} as follows:
\begin{align}
\label{ap:eq19}
       \mathbf{F}_k=
       \left[\begin{array}{llll}
		1 & 0 & 1 & 0  \\
		0 & 1 & 0 & 1  \\
		0 & 0 & 1 & 0  \\
            0 & 0 & 0 & 1
	\end{array}\right] 
        \quad
        \mathbf{H}_k=
       \left[\begin{array}{llll}
		1 & 0 & 0 & 0  \\
		0 & 1 & 0 & 0  
	\end{array}\right] 
        \quad
\end{align}

\end{document}